\documentclass[10pt,twocolumn,letterpaper]{article}

\usepackage[pagenumbers]{cvpr}
\usepackage[table]{xcolor}
\usepackage{multirow}

\newcommand{\systemname}{3DTouch}

\usepackage{cuted}
\usepackage{currfile}
\usepackage{caption}

\usepackage{colortbl}
\usepackage{pifont}
\usepackage{tabularx}
\usepackage{booktabs}
\usepackage{array}

\definecolor{fullrowgray}{RGB}{240,240,240}
\definecolor{fullrowblue}{RGB}{225,239,250}

\newcommand{\xmark}{\ding{56}}
\newsavebox{\ablationHeader}

\definecolor{cvprblue}{rgb}{0.21,0.49,0.74}
\definecolor{visionrow}{HTML}{F8F8F8}
\definecolor{bluerow}{HTML}{F3F8FC}
\definecolor{touchrow}{HTML}{F3F8FC}
\definecolor{oursrow}{HTML}{F7F4FA}

\usepackage[pagebackref,breaklinks,colorlinks,allcolors=cvprblue]{hyperref}

\title{
3DTouch: A Tactile-aware VLA Model with 3D Geometric Relations for Dexterous Manipulation
}

\author{
Xin Zhang$^{*}$ \quad
Shiqi Zhang$^{*}$ \quad
Siyuan Cao \quad
Juncheng Cai \quad
Yedong Shen \quad
Yuxuan Gao \\[3pt]
Dianlong Wen \quad
Kaixue Long \quad
Lu Zhang \quad
Yao Li$^{\dagger}$ \quad
Jiajun Deng \quad
Yanyong Zhang$^{\dagger}$ \\[5pt]
University of Science and Technology of China \\[3pt]
{\small $^{*}$Equal contribution.
\quad $^{\dagger}$Corresponding authors.}
}

\begin{document}

\maketitle

\begin{abstract}

In contact-rich dexterous manipulation, occlusion by the fingers and interacting objects obscures local geometry, preventing visual observations from fully capturing contact locations. Visually similar scenes may also differ subtly in the relative positions of interacting objects, leading to different contact states. We propose \textbf{3DTouch}, a tactile-geometric conditioning approach for vision-language-action (VLA) models. 3DTouch constructs a shared 3D visuo-tactile representation by expressing RGB-D scene points and dense five-finger tactile measurements in the robot-base frame, preserving each taxel's spatial location and force vector. Hand--object and object--object relations derived from hand landmarks and pose-aligned reconstructed surfaces explicitly describe the relative arrangement of interacting entities. Together, these representations condition the VLA action expert on spatially localized contact forces and relative surface geometry, providing explicit cues for contact adjustment and object alignment. We evaluate 3DTouch on five real-world contact-rich dexterous manipulation tasks using a UR7e robot arm equipped with an XHand dexterous hand. 3DTouch improves the average success rate by 42 percentage points over the corresponding VLA baseline and by 20 percentage points over the strongest evaluated alternative. Ablation studies reveal the complementary benefits of scene geometry and spatial tactile feedback, with further gains from shared spatial grounding and explicit geometric relations.

\end{abstract}

\section{Introduction}
\label{sec:intro}

\begin{figure*}[t]
    \centering
    \includegraphics[width=\textwidth]{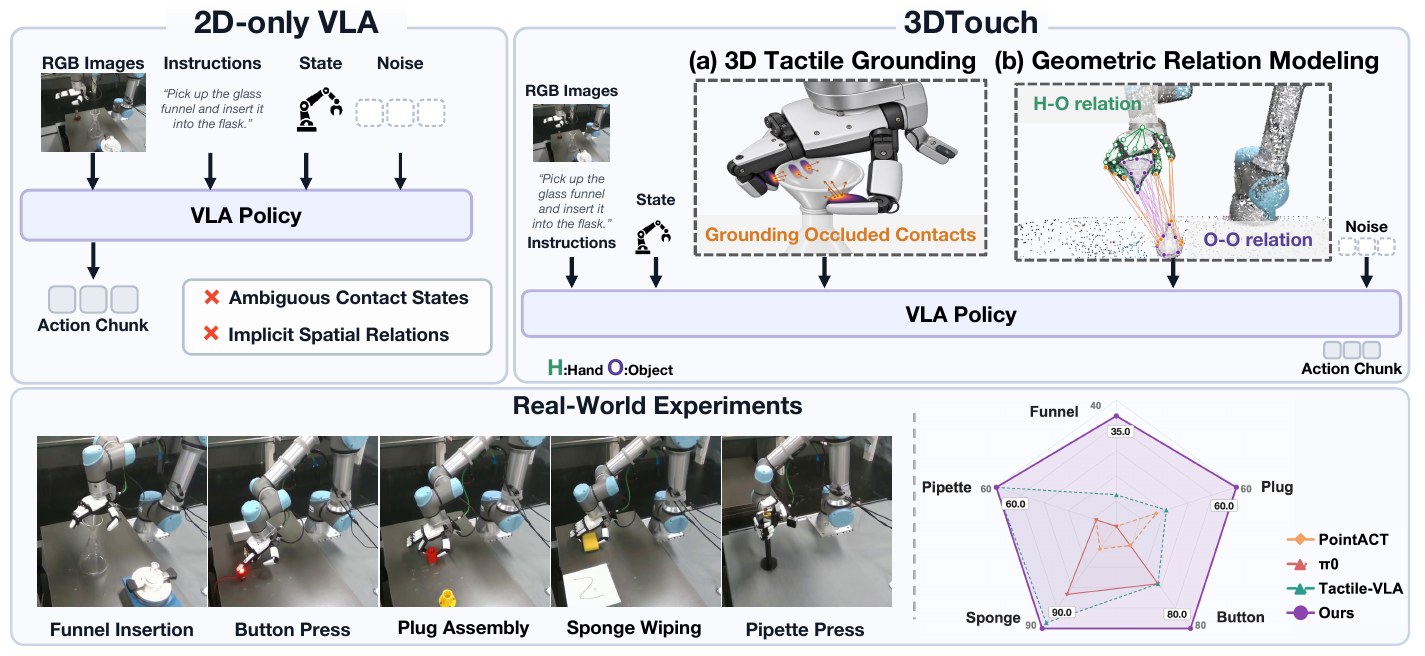}
    \caption{
\textbf{Overview of 3DTouch.}
Compared with the 2D-only VLA shown on the left, 3DTouch augments a pretrained vision-language-action policy with scene geometry, spatially grounded tactile feedback, and explicit geometric relations.
Scene points and tactile measurements are represented in a shared robot-base coordinate frame, while hand--object and object--object relations describe task-relevant geometric configurations.
These complementary representations jointly condition action generation, with the bottom panels illustrating five real-world dexterous manipulation tasks and their success-rate comparisons against baselines.
}
\label{fig:overview}
\end{figure*}

Vision-Language-Action (VLA) models have demonstrated promising capabilities in general robotic manipulation through large-scale pretraining on robot manipulation data  ~\cite{kim2025openvla,black2025pi0,black2025pi05}.
However, contact-rich dexterous manipulation remains challenging, as small deviations in object alignment or contact conditions can lead to task failure. Similar visual observations may correspond to different physical interaction states, requiring the model to refine actions according to additional contact information. Moreover, visual observations alone usually lack critical local manipulation geometry, especially under occlusion which commonly occurs in dexterous manipulation. These limitations show the necessity for fusing complementary geometric and tactile information into VLA models to support precise action control.

Recent studies have explored incorporating 3D spatial information and tactile feedback into robot policies. 
3D-aware VLA methods mainly adopt two schemes to integrate spatial information, using 3D positional encoding to fuse depth with visual features
~\cite{qu2025spatialvla},
or conditioning the action expert with separately encoded geometric features extracted from 3D reconstruction
~\cite{zhang2025spatial}
or observed point clouds
~\cite{chen2026pointact}.
Additionally, tactile-aware VLA methods incorporate tactile feedback into action generation by encoding tactile data as additional input tokens alongside visual-language tokens~\cite{huang2025tactilevla,zhang2025vtla}. 
3D visuo-tactile methods further represent visual observations and gripper tactile measurements as 3D points for visuo-tactile fusion, preserving their spatial relationships~\cite{huang2025vitac}.

Despite these advances, two key limitations remain for contact-rich dexterous manipulation. First, although 3D visuo-tactile methods have explored unified visual-tactile 3D representations for policy learning, how to effectively encode spatially grounded multi-finger tactile feedback for pretrained VLA models remains underexplored. 
In particular, without explicitly associating multi-finger taxel force measurements with their corresponding 3D locations in scenes, the model may struggle to relate local contact conditions of the hand to surrounding objects. 
Second, downsampled scene point clouds are often too coarse and heavily occluded during five-finger grasping to capture small interaction clearances. Even when complete object surfaces are tracked, implicitly inferring fine-grained hand--object and object--object alignment without explicit relational descriptors remains challenging.
These limitations motivate jointly modeling spatially grounded tactile feedback and explicit interaction relations to support precise  dexterous manipulation.

To address these challenges, we propose \textbf{3DTouch}, a 3D conditioning approach to enhance pretrained VLAs with spatially encoded tactile representations and explicit interaction relations for contact-rich dexterous manipulation. As shown in Fig.~\ref{fig:overview}, 3DTouch consists of two complementary modules. \textit{3D Tactile Grounding} (3DTG) represents RGB-D scene point clouds and dense five-finger tactile measurements in a shared robot-base coordinate system. By encoding each taxel's 3D position alongside force vectors, 3DTG provides tactile tokens spatially aligned with scene tokens. 
\textit{Geometric Relation Modeling} (GRM) further leverages kinematic hand landmarks and tracked 3D object surface points in the robot base frame to construct hand--object and object--object relation tokens, explicitly encoding relative displacements and distances between interacting entities. Together, spatially grounded tactile feedback supports
contact-aware motion refinement under visual ambiguity,
while explicit interaction geometry facilitates more precise
hand--object positioning and object--object alignment.
The resulting scene, tactile, and relation tokens jointly enhance VLA models for dexterous action generation.

We evaluate 3DTouch on five real-world contact-rich dexterous manipulation tasks using a UR7e robot arm equipped with an XHand dexterous hand. 3DTouch improves the average success rate by 42 percentage points over the corresponding VLA baseline and 20 percentage points over the strongest tactile-aware VLA. Ablation studies demonstrate the complementary benefits of scene geometry and spatial tactile feedback, with shared spatial grounding improving the average success rate by 11 percentage points over hand-local tactile encoding. Explicit interaction geometry further improves the average success rate by 12 percentage points, showing improvements in target localization and grasp stability during dexterous manipulation.

Our main contributions are summarized as follows:

\begin{itemize}
    \item We propose 3D Tactile Grounding (3DTG), which encodes dense five-finger tactile measurements and RGB-D scene geometry in a shared 3D coordinate system, providing spatially localized tactile feedback.

    \item We introduce Geometric Relation Modeling (GRM), which constructs structured hand--object and object--object representations from hand landmarks and 3D object geometry, explicitly encoding task-relevant relative displacements and distances for dexterous  action control.

    \item Extensive experiments demonstrate the effectiveness of 3DTouch on real-world dexterous manipulation tasks, with ablations validating the benefits of multimodal spatial grounding and explicit geometric relations, particularly with stronger target localization and grasp stability.

\end{itemize}

\section{Related Work}
\label{sec:related}

\textbf{Vision-Language-Action Models.} VLA models adapt pretrained vision-language representations for robotic control. RT-2 discretizes actions into tokens~\cite{rt2}, and OpenVLA learns from Open X-Embodiment demonstrations~\cite{kim2025openvla}. Moving beyond discrete action tokens, diffusion Transformers originally explored in image synthesis~\cite{peebles2023dit,shen2026hidit} have been adapted to robotic control~\cite{liu2025rdt}. \(\pi_0\) introduces a flow-matching action expert for continuous action generation~\cite{black2025pi0}, while \(\pi_{0.5}\) uses heterogeneous co-training for open-world generalization~\cite{black2025pi05}. To accelerate action generation, recent drifting-based policies enable native one-step inference~\cite{gao2026dbpo,gao2026driftingvla}. However, conventional VLA models primarily use vision and language, sometimes supplemented by proprioception, and lack direct measurements of local contact forces; contact states can remain ambiguous under visual occlusion. We extend \(\pi_0\) with tactile and geometric conditioning while retaining its vision-language input pathway and flow-matching objective.

\noindent\textbf{3D-Aware Manipulation Policies.} RVT processes rendered scene views~\cite{goyal2023rvt}, while 3D Diffuser Actor and DP3 condition action diffusion on 3D representations~\cite{ke2025diffactor,ze2024dp3}. Transferring pretrained 2D knowledge to point clouds, as explored in open-vocabulary 3D scene understanding~\cite{peng2023openscene,zhang2025pgov3d}, has also been applied to robotic manipulation, with Lift3D employing positional mapping~\cite{jia2025lift3d}. Within VLAs, SpatialVLA introduces spatial position encoding and adaptive action discretization~\cite{qu2025spatialvla}, PointACT employs multi-scale point--action attention~\cite{chen2026pointact}, while FALCON and GEAR-VLA leverage spatial foundation models to enhance action generation~\cite{zhang2025spatial,zhang2026gearvla}. We complement these approaches with tactile feedback and explicit surface relations computed using object shape priors aligned by estimated poses.

\noindent\textbf{Tactile-Aware Policies.} Early work uses vision and touch for grasp adjustment~\cite{calandra2018more}. Robot Synesthesia represents binary contact signals as 3D points~\cite{yuan2024robotsynesthesia}, while TARS learns visuo-tactile affordances through a unified point-cloud representation~\cite{wu2025tars}. 3D-ViTac associates scalar tactile readings with their 3D locations and jointly encodes visual and tactile points~\cite{huang2025vitac}. ForceVLA incorporates six-axis force feedback through a force-aware mixture-of-experts~\cite{yu2025forcevla}, while Tactile-VLA combines tactile feedback with hybrid position--force control and tactile-informed reasoning~\cite{huang2025tactilevla}. FAVLA and ReTouch further exploit high-frequency contact feedback for reactive action generation~\cite{li2026favla,zhang2026retouch}. OmniVTLA aligns tactile representations with visual and linguistic semantics through dual-path tactile encoding~\cite{cheng2025omnivtla}. MLA and DreamTacVLA establish spatial correspondences by projecting tactile sensor locations into image space for contrastive feature alignment, combined with future sensory prediction~\cite{liu2025mla,ye2025dreamtacvla}. In contrast, our 3D Tactile Grounding (3DTG) explicitly represents both the 3D location and force vector of each taxel across multiple fingers in a shared metric frame with scene geometry. Unlike 3D-ViTac's joint point-cloud encoding, 3DTG separately encodes scene geometry and finger-wise force distributions to accommodate their distinct spatial structures.

\noindent\textbf{Relational Geometry for Manipulation.} ReKep formulates manipulation tasks as relational keypoint constraints for trajectory optimization~\cite{huang2024rekep}, while OmniManip leverages reconstructed object meshes and estimated poses to define spatial constraints between object-centric interaction primitives~\cite{pan2025omnimanip}. TriRelVLA models entity-level interactions through a task-conditioned graph that combines learned object and hand representations with relative positions, poses, and distances~\cite{zhou2026trirelvla}. Rather than prescribing desired geometric constraints or modeling relations at the entity level, our Geometric Relation Modeling (GRM) explicitly characterizes current interaction geometry at the surface level, computing relative displacements and distances between kinematic hand landmarks and nearby object surfaces, and between representative regions of interacting objects. These fine-grained geometric relations complement spatially grounded multi-finger tactile feedback for contact-rich dexterous manipulation.
\section{Method}
\label{sec:method}

In this section, we present the proposed 3DTouch framework. Sec.~\ref{sec:overview} provides an overview of the approach. Sec.~\ref{sec:spatial_encoding} introduces 3D Tactile Grounding (3DTG), which represents scene geometry and tactile feedback in a shared 3D coordinate frame. Sec.~\ref{sec:interaction_geometry} describes Geometric Relation Modeling (GRM), which constructs hand--object and object--object geometric relations. Finally, Sec.~\ref{sec:action_prediction} details how these representations condition the pretrained VLA action expert and how the model is trained.

\subsection{Overview}
\label{sec:overview}

\begin{figure*}[t]
    \centering
    \includegraphics[width=\textwidth]{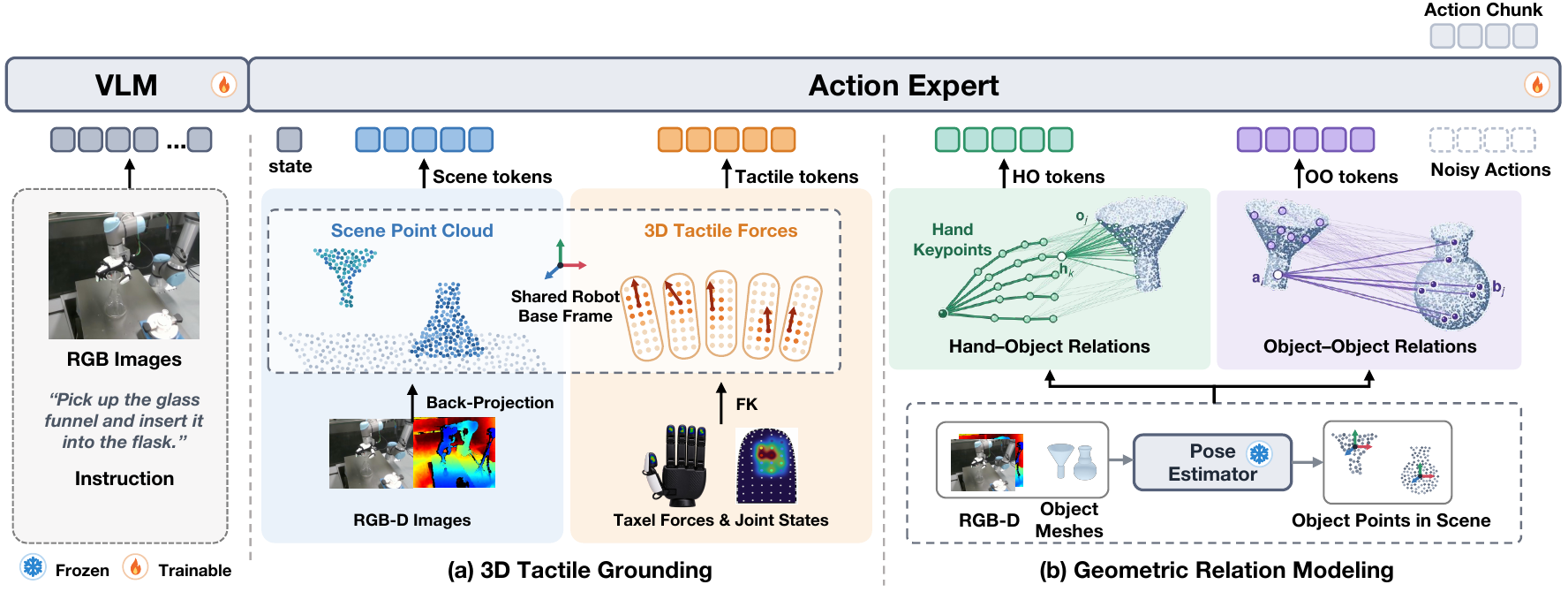}
    \caption{
\textbf{Pipeline of 3DTouch.}
(a) 3D Tactile Grounding (3DTG) back-projects RGB-D observations into scene point clouds and transforms taxel positions and force vectors into a shared robot-base frame using forward kinematics (FK).
Separate encoders produce scene tokens and finger-level tactile tokens.
(b) Geometric Relation Modeling (GRM) aligns reference object meshes using estimated poses and encodes relative displacements and distances between hand landmarks and nearby object surface points, and between representative regions of interacting objects, into hand--object (HO) and object--object (OO) tokens.
The scene, tactile, and relation tokens jointly condition the action expert, while the original vision-language pathway and flow-matching objective are preserved.
}
    \label{fig:framework}
\end{figure*}
We propose a method for contact-rich dexterous manipulation that incorporates two complementary types of 3D information into a pretrained vision-language-action (VLA) model. These include (i) scene point clouds and spatially grounded tactile observations, and (ii) explicit interaction geometry among task-relevant entities. As illustrated in Fig.~\ref{fig:framework}, our method consists of two modules: \textit{3D Tactile Grounding} and \textit{Geometric Relation Modeling}. The \textit{3D Tactile Grounding} module transforms RGB-D observations into scene point clouds in the robot base frame and extracts local and global scene features. In the same coordinate frame, it encodes the spatial positions and force measurements of tactile taxels on each finger into tactile tokens. The \textit{Geometric Relation Modeling} module directly models relative geometric relations between entities, rather than relying solely on their individual geometric features. It constructs Hand-Object (HO) and Object-Object (OO) relations, which are encoded into HO and OO tokens, respectively. The resulting scene, tactile, and interaction tokens are concatenated to provide complementary geometric and contact information for action prediction.

\subsection{3D Tactile Grounding}
\label{sec:spatial_encoding}

\noindent\textbf{Scene Point-Cloud Encoding.}
We construct scene point clouds from RGB-D observations by back-projecting depth measurements and transforming the resulting points into the robot base frame. After workspace cropping, voxel downsampling, and fixed-size sampling, we obtain $\mathcal P_t=\{(\mathbf p_i,\mathbf c_i,v_i)\}_{i=1}^{N}$ with $N=4096$, where $\mathbf p_i$, $\mathbf c_i$, and $v_i$ denote the 3D position, RGB color, and color-validity indicator, respectively.

To capture scene geometry at multiple spatial scales, we adopt a PointNet++~\cite{qi2017pointnetplusplus}-style
hierarchical set-abstraction encoder. Two successive FPS-KNN stages progressively sample 256 and 32 centers, with 24 and 16 neighbors per center, respectively. The first stage aggregates relative coordinates, RGB features, and color-validity indicators through shared MLPs and mean-max pooling. The second stage further aggregates first-stage features and relative center coordinates, with an MLP encoding of each center's absolute position in the robot base frame added to the resulting local token, alongside a modality embedding. In parallel, a global branch encodes all input points using centroid-relative coordinates and RGB features, followed by global average pooling. An MLP encoding of the scene centroid's absolute position and a modality embedding are then added to obtain the global token. The local and global features are concatenated to form the scene representation:
\begin{equation}
\mathbf Z_t^{\mathrm{pc}}
=
\operatorname{Concat}
\left(
\mathbf Z_t^{\mathrm{local}},
\mathbf z_t^{\mathrm{global}}
\right)
\in\mathbb R^{N_s\times d},
\end{equation}
where $N_s$ denotes the number of scene tokens and $d$ is the token dimension.

\noindent\textbf{Spatial Tactile Encoding.}
Beyond force magnitude and direction, the spatial location of tactile measurements provides important cues for interpreting local contact states. We therefore associate each tactile sensing element (taxel) with its 3D position and corresponding tactile force vector in the robot base frame.

Our five-finger dexterous hand contains 120 taxels per finger. Each taxel has a known, fixed position in its finger-local coordinate frame. Using the current joint states and robot forward kinematics, we transform these positions and the corresponding 3D tactile force vectors into the robot base frame, aligning their spatial reference with the scene point cloud.

Each taxel is represented by a six-dimensional feature \(\mathbf x_{t,f,j}=[\mathbf p_{t,f,j};\tilde{\mathbf f}_{t,f,j}]\in\mathbb R^6\), comprising its 3D position in the robot base frame and a normalized tactile force vector. The taxel features of each finger are concatenated and encoded by a shared MLP $\phi_{\mathrm{tac}}$:
\begin{equation}
\begin{aligned}
\mathbf z_{t,f}^{\mathrm{tac}}
&=
\phi_{\mathrm{tac}}
\left(
\operatorname{Concat}_{j=1}^{120}
\mathbf x_{t,f,j}
\right),\\
\mathbf Z_t^{\mathrm{tac}}
&=
\operatorname{Stack}_{f=1}^{5}
\mathbf z_{t,f}^{\mathrm{tac}}
\in\mathbb R^{5\times d}.
\end{aligned}
\end{equation}

We encode scene points and tactile taxels separately to accommodate their different sampling requirements: FPS-based scene encoding prioritizes broad geometric coverage, whereas tactile encoding must preserve spatially concentrated force measurements across each finger, which could otherwise be underrepresented by the same sampling strategy. Both encoders incorporate absolute positions in the shared robot-base frame, allowing their spatially referenced features to be jointly processed by the action expert.

\subsection{Geometric Relation Modeling}
\label{sec:interaction_geometry}

Although scene point clouds capture global geometry and spatial tactile observations provide local contact feedback, their encoded features do not explicitly characterize geometric relations between interacting entities. Such relations are particularly important during precise approach, alignment, and contact adjustment. We therefore construct explicit hand-object (HO) and object-object (OO) relations to complement the sensory representations.

\noindent\textbf{Geometry Construction.}

We obtain 17 finger keypoints from the current joint states through forward kinematics. Reference object meshes are reconstructed offline from unobstructed observations using SAM 3D~\cite{chen2026sam3d}, from which object surface points are sampled for subsequent geometric relation modeling.  During offline training-data preparation, SAM2~\cite{ravi2025sam2} generates object segmentation masks, while RGBTrack~\cite{guo2025rgbtrack} provides frame-wise 6D pose pseudo-labels to supervise the pose estimator.

\begin{figure}[t]
    \centering
    \includegraphics[width=0.98\linewidth]{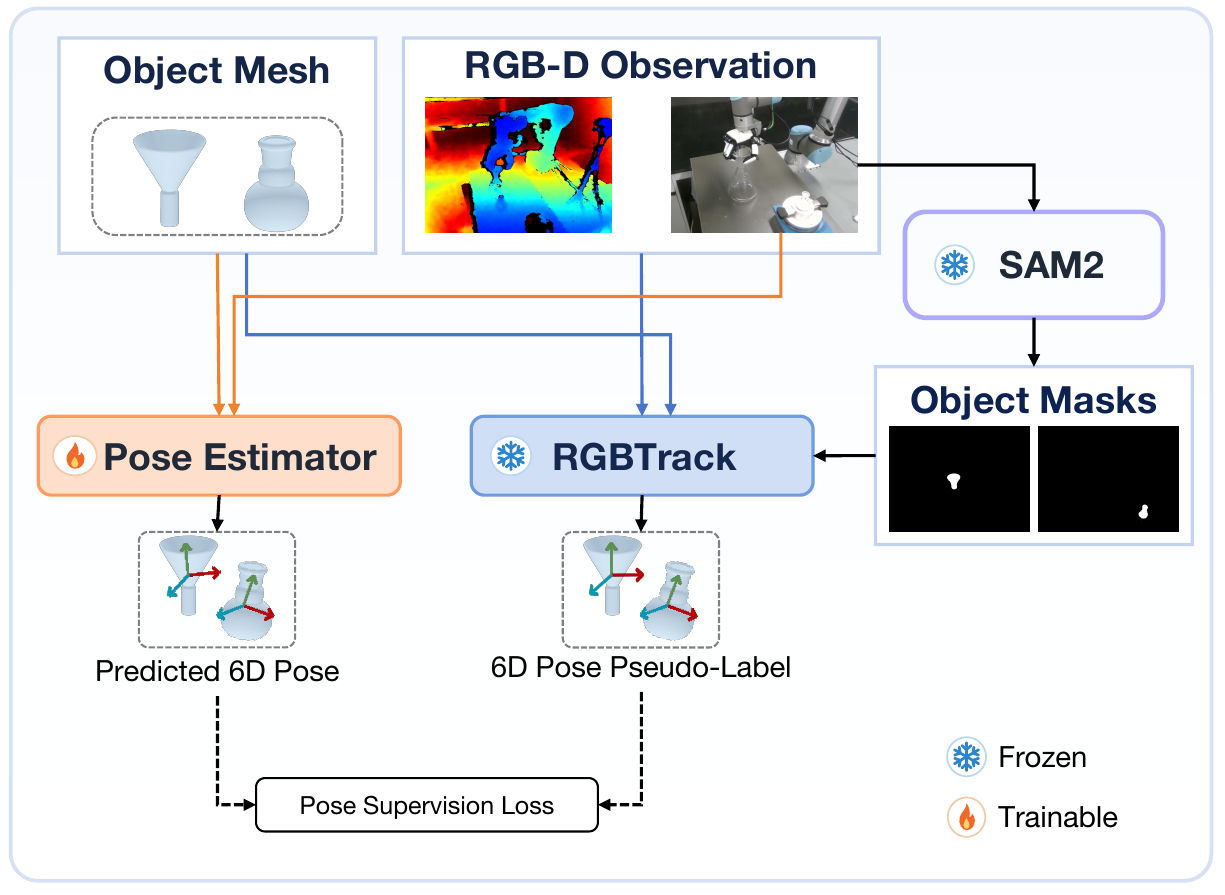}
    \caption{
\textbf{Pose Estimator.} SAM2 generates object segmentation masks, which RGBTrack uses together with RGB-D observations and reconstructed reference meshes to obtain offline 6D pose annotations. These annotations supervise a lightweight pose estimator that predicts object poses directly from RGB-D observations and sampled reference surface points. 
}
    \label{fig:fig3}
\end{figure}

\noindent\textbf{Pose Estimator.}
To avoid running the SAM2--RGBTrack pipeline online, we train a lightweight pose estimator, as illustrated in Fig.~\ref{fig:fig3}. It predicts each object's 6D pose in the robot-base frame from the current RGB-D observation and sampled reference surface points. Separate lightweight CNNs encode RGB and depth observations, while the reference points are embedded and pooled into an object query that attends to the visual features through cross-attention. We supervise the estimator with offline pose pseudo-labels using translation, rotation, and surface-point alignment losses. The alignment term compares reference points transformed by the predicted and pseudo-labeled poses. The predicted poses map the reference points into the robot-base frame, where they are used alongside hand keypoints to construct hand--object and object--object relations.

\noindent\textbf{Hand-Object Relations.}
We characterize local hand-object geometry by relating each hand keypoint to its $K$ nearest object surface points. Given a hand keypoint $\mathbf h_k$ and a neighboring object point $\mathbf o_j$, their relative geometric feature is defined as
\begin{equation}
\mathbf r_{k,j}^{\mathrm{HO}}
=
\left[
\mathbf o_j-\mathbf h_k;
\left\|\mathbf o_j-\mathbf h_k\right\|_2
\right]
\in\mathbb R^4,
\end{equation}
encoding relative displacement and Euclidean distance. We set \(K=16\) and project each 4D relation feature into a \(d\)-dimensional embedding. For each hand keypoint, mean and max pooling are applied across its 16 neighboring embeddings, followed by a linear projection to obtain a single landmark-level relation feature. We then group the landmark features by finger and apply a second mean-max pooling followed by a linear projection. By incorporating finger identity and relation-type embeddings, we obtain one HO token per finger, yielding \(\mathbf Z_t^{\mathrm{HO}}\in\mathbb R^{5\times d}\).

\noindent\textbf{Object-Object Relations.}
Hand-object geometry alone does not explicitly describe the relative alignment of interacting objects, which is essential for tasks such as insertion. For a task-relevant object pair, we first select \(M=8\) anchors per object using farthest point sampling (FPS), and assign each surface point to its nearest anchor. The mean position of each resulting region defines a representative geometric center, denoted by \(\{\mathbf a_i\}_{i=1}^{M}\) and \(\{\mathbf b_j\}_{j=1}^{M}\). Their pairwise geometric relations are defined as
\begin{equation}
\mathbf r_{i,j}^{\mathrm{OO}}
=
\left[
\mathbf b_j-\mathbf a_i;
\left\|\mathbf b_j-\mathbf a_i\right\|_2
\right]
\in\mathbb R^4.
\end{equation}
A shared two-layer MLP encodes the \(M^2=64\) pairwise relations into \(d\)-dimensional features. We then employ \(Q=5\) learnable queries, each attending to all 64 relation features to produce a weighted aggregation, yielding \(\mathbf Z_t^{\mathrm{OO}}\in\mathbb R^{5\times d}\). The HO and OO tokens are concatenated into $\mathbf Z_t^{\mathrm{rel}}=\operatorname{Concat}(\mathbf Z_t^{\mathrm{HO}},\mathbf Z_t^{\mathrm{OO}})\in\mathbb R^{10\times d}$. 

\subsection{Action Prediction and Training}
\label{sec:action_prediction}

Our method incorporates the proposed 3D representations into pretrained VLA policies. The scene, tactile, and interaction tokens are concatenated and mapped into the Action Expert's embedding space through a learnable linear projection:
\begin{equation}
\begin{aligned}
\mathbf Z_t^{\mathrm{3D}}
&=\operatorname{Concat}\left(
\mathbf Z_t^{\mathrm{pc}},
\mathbf Z_t^{\mathrm{tac}},
\mathbf Z_t^{\mathrm{rel}}
\right)
\in\mathbb R^{(N_s+15)\times d},\\
\widehat{\mathbf Z}_t^{\mathrm{3D}}
&=\operatorname{Proj}_{\theta}
\left(\mathbf Z_t^{\mathrm{3D}}\right).
\end{aligned}
\end{equation}
The projected tokens are inserted into the Action Expert's observation suffix before the noisy action tokens, while the original vision-language prefix is preserved.

We retain the original flow-matching training objective without introducing additional losses. During inference, the 3D representations are computed once per observation and reused across denoising steps.
\section{Experiments}

\begin{figure*}[t]
    \centering
    \includegraphics[width=0.95\textwidth]{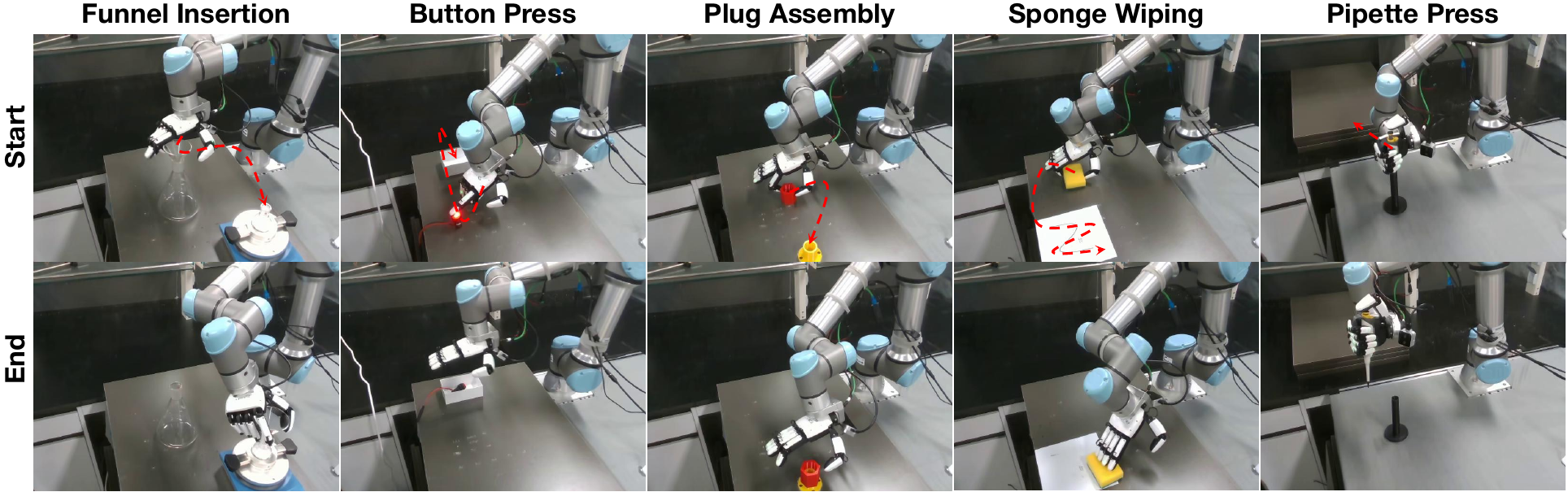}
    \caption{Overview of five contact-rich manipulation tasks performed on the UR7e robot equipped with an XHand dexterous hand. Each column shows the start and end states of a task, with red dashed arrows indicating the intended manipulation motions.}
    \label{fig:experiments}
\end{figure*}

\begin{table*}[t]
\centering
\caption{
Success rates (\%) on real-world contact-rich dexterous manipulation tasks.
Each method is evaluated over 20 real-robot rollouts per task.
Baselines are grouped into point-cloud policy (w or w/o tactile feedback), vision-based policy, and tactile-augmented policy.
}
\label{tab:main_results}

\setlength{\tabcolsep}{4pt}
\renewcommand{\arraystretch}{1.15}
\small

\begin{tabularx}{\textwidth}{
@{}
>{\centering\arraybackslash}p{2.5cm}
>{\raggedright\arraybackslash}p{3.3cm}
*{6}{>{\centering\arraybackslash}X}
@{}
}
\toprule

\raisebox{1.2ex}{\textbf{Category}}
& \raisebox{1.2ex}{\textbf{Method}}
& \shortstack{\textbf{Funnel}\\\textbf{Insertion}}
& \shortstack{\textbf{Plug}\\\textbf{Assembly}}
& \shortstack{\textbf{Button}\\\textbf{Press}}
& \shortstack{\textbf{Sponge}\\\textbf{Wiping}}
& \shortstack{\textbf{Pipette}\\\textbf{Press}}
& \raisebox{1.2ex}{\textbf{Avg.} $\uparrow$} \\

\midrule

\multirow{2}{2.5cm}{\centering Vision-based\\Policy}
& $\pi_{0}$~\cite{black2025pi0}
& 0.0 & 0.0 & 45.0 & 60.0 & 10.0 & 23.0 \\

& $\pi_{0.5}$~\cite{black2025pi05}
& 15.0 & 50.0 & 75.0 & 60.0 & 20.0 & 44.0 \\

\midrule

\multirow{2}{2.5cm}{\centering Point-cloud\\Policy}
& Simple DP3 + tactile~\cite{ze2024dp3}
& 10.0 & 0.0 & 10.0 & 20.0 & 10.0 & 10.0 \\

& PointACT~\cite{chen2026pointact}
& 0.0 & 20.0 & 15.0 & 20.0 & 10.0 & 13.0 \\

\midrule

\multirow{3}{2.5cm}{\centering Tactile-augmented\\Policy}
& $\pi_{0}$ + tactile
& 0.0 & 15.0 & 45.0 & 70.0 & 25.0 & 31.0 \\

& $\pi_{0.5}$ + tactile
& 20.0 & 55.0 & 30.0 & 70.0 & 40.0 & 43.0 \\

& Tactile-VLA~\cite{huang2025tactilevla}
& 10.0 & 25.0 & 45.0 & 85.0 & \textbf{60.0} & 45.0 \\

\midrule
\rowcolor{bluerow}

\textbf{Ours}
& \textbf{\systemname{}}
& \textbf{35.0}
& \textbf{60.0}
& \textbf{80.0}
& \textbf{90.0}
& \textbf{60.0}
& \textbf{65.0} \\

\bottomrule
\end{tabularx}
\end{table*}

\begin{table}[t]
\centering
\caption{
Robustness to target pose variations on
Funnel Insertion and Plug Assembly.
Trans., Rot., and Both denote translation,
rotation, and their combination, respectively.
Success rates (\%) are reported over
20 real-robot trials per task and setting.
}
\label{tab:target_pose_variation}

\setlength{\tabcolsep}{2.5pt}
\renewcommand{\arraystretch}{1.15}
\small

\begin{tabularx}{\columnwidth}
{@{}l*{6}{>{\centering\arraybackslash}X}@{}}
\toprule

\multirow{2}{*}{\textbf{Method}}
& \multicolumn{3}{c}{\textbf{Funnel Insertion}}
& \multicolumn{3}{c}{\textbf{Plug Assembly}} \\

\cmidrule(lr){2-4}
\cmidrule(l){5-7}

& \textbf{Trans.}
& \textbf{Rot.}
& \textbf{Both}
& \textbf{Trans.}
& \textbf{Rot.}
& \textbf{Both} \\

\midrule

$\pi_{0.5}$ + tactile
& 10.0 & 0.0 & 0.0
& \textbf{55.0} & 50.0 & 30.0 \\

\rowcolor{bluerow}

\textbf{3DTouch}
& \textbf{30.0} & \textbf{30.0} & \textbf{30.0}
& \textbf{55.0} & \textbf{60.0} & \textbf{50.0} \\

\bottomrule
\end{tabularx}
\end{table}

\begin{table*}[t]
\centering
\caption{
Ablation study of 3DTouch across five real-world
contact-rich dexterous manipulation tasks.
We investigate the individual and combined contributions
of spatial tactile sensing, scene point clouds, and
interaction geometry, as well as tactile coordinate alignment.
Global and Local denote robot-base-frame and hand-local
tactile representations, respectively.
Success rates (\%) are reported. Values in
red parentheses indicate changes in average success rate (percentage points) relative to the full model.
}
\label{tab:ablation_results}

\setlength{\tabcolsep}{1.5pt}
\renewcommand{\arraystretch}{1.14}
\renewcommand{\xmark}{\ensuremath{\times}}
\fontsize{8.5}{10.1}\selectfont

\renewcommand{\tabularxcolumn}[1]{m{#1}}

\sbox{\ablationHeader}{%
  \shortstack{\textbf{Button}\\\textbf{Press}}%
}

\begin{tabularx}{\textwidth}{
@{}
l
*{8}{>{\centering\arraybackslash}X}
c
@{}
}
\toprule

\multirow[c]{2}{*}[
  \dimexpr
    (\ht\strutbox+\dp\strutbox
     -\ht\ablationHeader-\dp\ablationHeader)/2
    -\belowrulesep
  \relax
]{\textbf{Variant}}
& \multicolumn{3}{c}{\textbf{Input Configuration}}
& \multicolumn{6}{c}{\textbf{Task Success Rate (\%)}} \\

\cmidrule(lr){2-4}
\cmidrule(l){5-10}

& \textbf{Tactile}
& \shortstack{\textbf{Point}\\\textbf{Cloud}}
& \shortstack{\textbf{Interaction}\\\textbf{Geometry}}
& \shortstack{\textbf{Funnel}\\\textbf{Insertion}}
& \shortstack{\textbf{Plug}\\\textbf{Assembly}}
& \shortstack{\textbf{Button}\\\textbf{Press}}
& \shortstack{\textbf{Sponge}\\\textbf{Wiping}}
& \shortstack{\textbf{Pipette}\\\textbf{Press}}
& \textbf{Avg.} $\uparrow$ \\

\midrule

\rowcolor{bluerow}
\textbf{3DTouch (Full)}
& \textbf{Global}
& $\checkmark$
& $\checkmark$
& \textbf{35.0} & \textbf{60.0} & \textbf{80.0}
& \textbf{90.0} & \textbf{60.0}
& \textbf{65.0} \\

\midrule

w/o Geometric Relations
& Global & $\checkmark$ & \xmark
& 15.0 & \textbf{60.0} & 55.0 & 80.0 & 55.0
& \mbox{53.0\,{\scriptsize\textcolor{red}{($-12.0$)}}} \\

Local Tactile
& Local & $\checkmark$ & \xmark
& 15.0 & 40.0 & 25.0 & \textbf{90.0} & 40.0
& \mbox{42.0\,{\scriptsize\textcolor{red}{($-23.0$)}}} \\

\addlinespace[2pt]

Point Cloud Only
& \xmark & $\checkmark$ & \xmark
& 10.0 & 50.0 & 50.0 & 80.0 & 40.0
& \mbox{46.0\,{\scriptsize\textcolor{red}{($-19.0$)}}} \\

Tactile Only
& Global & \xmark & \xmark
& 10.0 & 50.0 & 30.0 & 60.0 & 25.0
& \mbox{35.0\,{\scriptsize\textcolor{red}{($-30.0$)}}} \\

Geometric Relations Only
& \xmark & \xmark & $\checkmark$
& 5.0 & 45.0 & 50.0 & 70.0 & 35.0
& \mbox{41.0\,{\scriptsize\textcolor{red}{($-24.0$)}}} \\

RGB Only
& \xmark & \xmark & \xmark
& 0.0 & 0.0 & 45.0 & 60.0 & 10.0
& \mbox{23.0\,{\scriptsize\textcolor{red}{($-42.0$)}}} \\

\bottomrule
\end{tabularx}
\end{table*}

\begin{figure}[t]
    \centering
    \includegraphics[width=\linewidth]{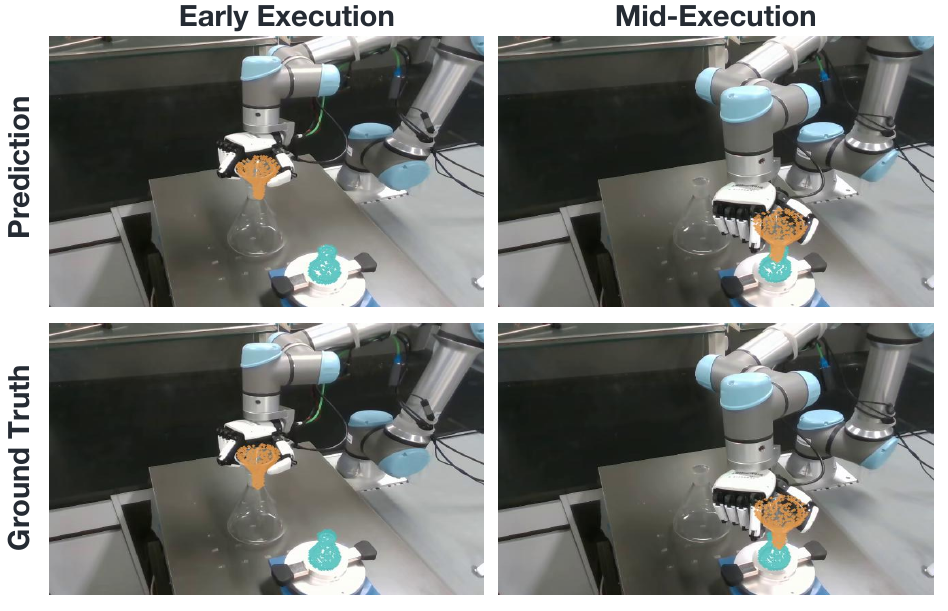}
    \caption{
\textbf{Qualitative Evaluation of Pose Estimation.}
Object surface points are transformed into the robot base frame using predicted poses (top) or reference poses from the offline SAM2--RGBTrack pipeline (bottom), and reprojected onto RGB images. Similar overlays demonstrate that our estimator produces object alignments consistent with the offline pose annotations.
    }
    \label{fig:fig4}
\end{figure}

\begin{figure}[t]
    \centering
    \includegraphics[width=\linewidth]{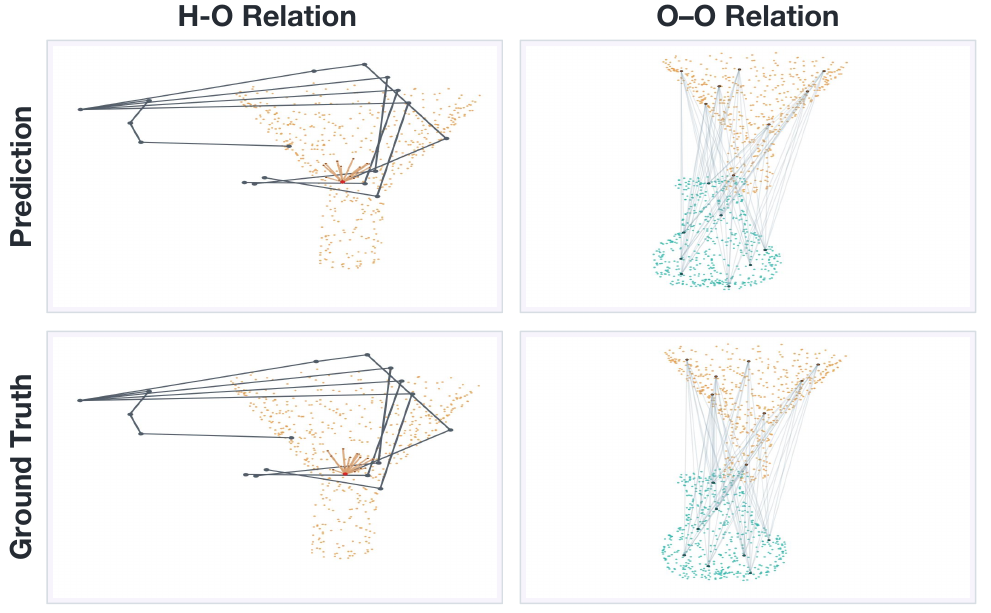}
    \caption{
    \textbf{Visualization of Geometric Relations.}
    Hand--object (HO) and object--object (OO) relations
    constructed using object poses predicted by our pose
    estimator (top) and reference poses obtained from the
    offline SAM2--RGBTrack pipeline (bottom).
    The similar spatial configurations illustrate the
    consistency of the resulting interaction geometry.
    }
    \label{fig:ho_oo}
\end{figure}

\subsection{Experimental Setup}

\noindent\textbf{Robot Platform.}
We evaluate 3DTouch on a real-world manipulation platform consisting of a
UR7e robot arm and an XHand dexterous hand. The XHand is equipped with dense
five-finger tactile sensing, with 120 three-dimensional force taxels per
finger. We use a front camera view for visual observations and
calibrated RGB-D sensing to obtain scene geometry. 

\noindent\textbf{Data and Tasks.}
We collect approximately 100 real-robot demonstrations for each of five contact-rich manipulation tasks: (1) \textit{Funnel Insertion}, inserting a funnel into the
opening of a flask; (2) \textit{Plug Assembly}, aligning and inserting a plug
into its corresponding socket; (3) \textit{	Button Press}, pressing
a button and completing a subsequent transfer operation; (4)
\textit{Sponge Wiping}, grasping a sponge and wiping handwritten marks from a
whiteboard; and (5) \textit{Pipette Press}, removing a pipette from its holder and pressing its plunger. The demonstrations include visual observations, task instructions, robot states, actions, and five-finger tactile measurements. 

\noindent\textbf{Evaluation Protocol and Metrics.}
We evaluate all methods on real-robot rollouts using binary task success, where a trial is considered successful only if the complete task objective is achieved.
We report the success rate (\%) for each task and the unweighted mean across all five tasks. All methods are evaluated over 20 real-robot trials per task, following the same task-specific success criteria and evaluation protocol.

\noindent\textbf{Baselines.}
We compare 3DTouch with seven baselines: $\pi_0$, $\pi_{0.5}$,
$\pi_0$ + tactile, $\pi_{0.5}$ + tactile, Tactile-VLA, Simple DP3 + tactile,
and PointACT. The $\pi_0$ and $\pi_{0.5}$ baselines are pretrained
vision-language-action policies fine-tuned on task-specific demonstrations. For the \(\pi_0\) + tactile and \(\pi_{0.5}\) + tactile baselines, we incorporate raw tactile measurements into the robot state inputs without explicit spatial encoding. Tactile-VLA serves as a tactile-aware VLA baseline, Simple DP3 + tactile as our tactile-augmented variant of the point-cloud diffusion policy, and PointACT as a 3D-aware VLA with multi-scale point–action interaction. We train all baselines on the same task-specific demonstrations, adapting the input modalities to each method's requirements.

\noindent\textbf{Implementation Details.}
We sample 4,096 points from RGB-D scene point clouds and retain tactile measurements from all 600 taxels. The sampled scene points and tactile measurements are encoded into 33 scene tokens and 5 tactile tokens, respectively.
The GRM module generates 5 hand–object and 5 object–object relation tokens. We implement 3DTouch using a pretrained \(\pi_0\) backbone with 18-dimensional joint actions (6 arm and 12 hand joints), executing 16 actions per inference.

\subsection{Main Results and Analysis}

\noindent{\textbf{{Overall Task Performance.}}}
Table~\ref{tab:main_results} summarizes the success rates on five real-world contact-rich dexterous manipulation tasks. 3DTouch achieves the highest average success rate of 65.0\%, outperforming Tactile-VLA, the strongest baseline on average, by 20.0 percentage points. 3DTouch achieves the highest success rate on four tasks and matches Tactile-VLA on \textit{Pipette Press}. The largest margin over the strongest task-specific baseline occurs on \textit{Funnel Insertion}, where 3DTouch achieves a 35.0\% success rate, exceeding the best baseline by 15.0 percentage points.

Two observations emerge. First, adding raw tactile measurements to pretrained VLA policies does not consistently improve task success. The effects vary substantially across tasks: tactile augmentation improves \(\pi_{0.5}\) on four tasks but reduces its success rate on \textit{Button Press} from 75.0\% to 30.0\%. Second, compared with Tactile-VLA, 3DTouch achieves larger gains on tasks involving precise geometric alignment and contact, improving success rates by 25.0, 35.0, and 35.0 percentage points on \textit{Funnel Insertion}, \textit{Plug Assembly}, and \textit{Button Press}, respectively. By contrast, the improvement is 5.0 percentage points on \textit{Sponge Wiping}, while both methods achieve the same success rate on \textit{Pipette Press}.

\noindent{\textbf{Visualization of Object Pose Estimation.}}
Fig.~\ref{fig:fig4} compares the object poses predicted by our lightweight estimator with reference poses obtained from the offline SAM2--RGBTrack pipeline at two stages of \textit{Funnel Insertion}. Object surface points are transformed into the robot base frame using the corresponding poses and then reprojected onto the RGB observations. The resulting overlays show similar alignment with the observed objects, indicating that our lightweight estimator produces geometrically consistent pose predictions relative to the offline annotations.

Fig.~\ref{fig:ho_oo} further compares the hand--object and object--object relations constructed from predicted object poses with those obtained using offline reference poses. The visualizations show similar spatial arrangements of hand keypoints, neighboring object surface points, and inter-object geometric connections, indicating that the lightweight pose estimator preserves the principal geometric structures used by GRM relative to the offline annotations.

\subsection{Robustness to Target Pose Variations}

We further compare the robustness of 3DTouch and \(\pi_{0.5}\) + tactile to target pose variations, including translation, rotation, and their combination, on \textit{Funnel Insertion} and \textit{Plug Assembly}, as shown in Table~\ref{tab:target_pose_variation}. We select this baseline because it achieves the highest success rates among all evaluated baselines on both tasks under the original evaluation settings. These slight pose variations test whether the policy can adapt to small changes in target position and orientation while maintaining the precise alignment and contact required for successful manipulation. Averaged across both tasks and all three perturbation settings, the success rate of 3DTouch decreases by 5.0 percentage points relative to the unperturbed setting, compared with 13.3 percentage points for \(\pi_{0.5}\) + tactile. This difference is particularly evident on \textit{Funnel Insertion}, where the tight geometric tolerance required for insertion makes even mild pose variations challenging: the success rate decreases by 5.0 percentage points for 3DTouch, compared with 16.7 percentage points for the baseline. These findings suggest that grounding action prediction in the current interaction geometry can help the policy accommodate mild target pose variations, particularly when precise alignment is critical.

\subsection{Ablation Studies} 
Table~\ref{tab:ablation_results} investigates the contributions of scene point clouds, spatial tactile sensing, and explicit interaction geometry. Starting from the RGB-only baseline (23.0\%), independently incorporating spatial tactile sensing, point clouds, or interaction geometry improves the average success rate to 35.0\%, 46.0\%, and 41.0\%, respectively. Combining point clouds with spatial tactile sensing further increases the average success rate to 53.0\%, outperforming point clouds alone by 7.0 percentage points and spatial tactile sensing alone by 18.0 percentage points. This suggests that scene geometry and spatially grounded tactile feedback provide complementary information for contact-rich manipulation.

Importantly, representing tactile taxel positions in the hand-local frame rather than the robot base frame reduces the average success rate from 53.0\% to 42.0\%. The largest degradations occur on \textit{Plug Assembly} (60.0\% to 40.0\%) and \textit{Button Press} (55.0\% to 25.0\%). Qualitative rollout observations further reveal more persistent unintended contacts with hand-local tactile coordinates, including sustained pressure against the flask during \textit{Funnel Insertion} and excessive button pressing during \textit{Button Press}. These results support the benefit of representing tactile taxel positions and scene geometry in a shared 3D coordinate frame, although the performance gains vary across tasks. 

Finally, adding explicit interaction geometry to the combined scene and tactile representations increases the average success rate from 53.0\% to 65.0\%, with the largest gains of 20.0 and 25.0 percentage points on \textit{Funnel Insertion} and \textit{Button Press}, respectively. These results suggest that explicit interaction geometry provides useful conditioning both when added to the original VLA inputs and when combined with scene geometry and spatial tactile feedback.

\section{Conclusion}
\label{sec:conclusion}

We introduced 3DTouch, a tactile-geometric conditioning approach that grounds tactile feedback in scene geometry and explicitly models interaction relations for contact-rich dexterous manipulation. Across five real-world tasks, 3DTouch achieves an average success rate of 65.0\%, outperforming the strongest evaluated baseline by 20.0 percentage points. Ablations reveal complementary benefits of scene geometry and spatial tactile feedback, with further gains from shared spatial grounding and explicit interaction geometry. 3DTouch also exhibits smaller performance degradation under mild target pose variations than a tactile-augmented VLA baseline. These findings highlight the importance of interpreting local contact forces in their spatial context and explicitly modeling the geometry of interacting entities. Future work will focus on reducing reliance on reference object geometry and pose estimation and extending 3DTouch to diverse robot embodiments.

{
    \small
    \bibliographystyle{ieeenat_fullname}
    \bibliography{main}
}

\end{document}